\documentclass[runningheads]{llncs}
\usepackage[mobile]{eccv}
\usepackage{eccvabbrv}

\usepackage[accsupp]{axessibility}
\usepackage{algorithm}
\usepackage{amsfonts}
\usepackage{amsmath}
\usepackage{booktabs}
\usepackage{caption}
\usepackage{easy-todo}
\usepackage{graphicx}
\usepackage{microtype}
\usepackage{multirow}
\usepackage{pifont}
\usepackage{siunitx}
\usepackage{url}
\usepackage{wrapfig}
\usepackage{xspace}
\usepackage[breaklinks,colorlinks,citecolor=eccvblue]{hyperref}
\usepackage[capitalize]{cleveref}

\DeclareMathOperator*{\argmax}{argmax}
\DeclareMathOperator*{\argmin}{argmin}

\newcommand{\method}{SHIC\xspace}

\title{\method: Shape-Image Correspondences with \texorpdfstring{\\}{} no Keypoint Supervision}
\titlerunning{SHIC}
\author{Aleksandar Shtedritski\and
Christian Rupprecht \and
Andrea Vedaldi}
\authorrunning{A.~Shtedritski et al.}
\institute{{Visual Geometry Group, University of Oxford}\\
\email{\{suny, chrisr, vedaldi\}@robots.ox.ac.uk} \\
\email{ \href{https://www.robots.ox.ac.uk/~vgg/research/shic/}{robots.ox.ac.uk/vgg/research/shic/}}}

\begin{document}
\maketitle
\begin{figure}[h]
\begin{center}
\includegraphics[width=0.87\textwidth]{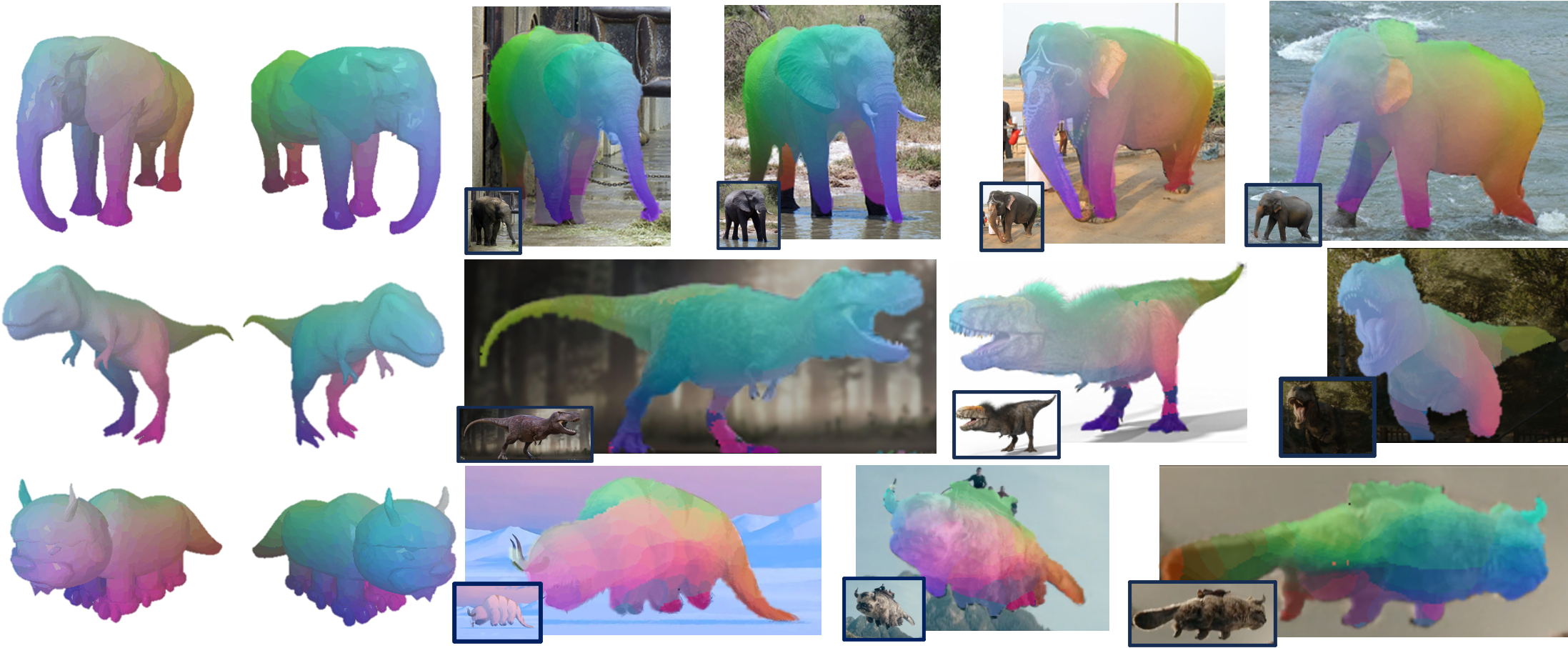}
\caption{\textbf{Unsupervised canonical maps.}
We show predictions from our \emph{fully unsupervised} method \method, which finds correspondences between a rigid 3D template and a natural image.
Correspondences are color-coded by assigning a distinct color to each template surface point.
Our approach is highly data-efficient; the elephant, T-Rex, and Appa models above are trained on only 2800, 480, and 180 images, respectively.
}%
\label{fig:teaser}
\end{center}\vspace{-1em}
\end{figure}
\begin{abstract}
Canonical surface mapping generalizes keypoint detection by assigning each pixel of an object to a corresponding point in a 3D template.
Popularised by DensePose for the analysis of humans, authors have since attempted to apply the concept to more categories, but with limited success due to the high cost of manual supervision.
In this work, we introduce \method, a method to learn canonical maps without manual supervision which achieves \emph{better} results than supervised methods for most categories.
Our idea is to leverage foundation computer vision models such as DINO and Stable Diffusion that are open-ended and thus possess excellent priors over natural categories.
\method reduces the problem of estimating image-to-template correspondences to predicting image-to-image correspondences using features from the foundation models.
The reduction works by matching images of the object to non-photorealistic renders of the template, which emulates the process of collecting manual annotations for this task.
These correspondences are then used to supervise high-quality canonical maps for any object of interest.
We also show that image generators can further improve the realism of the template views, which provide an additional source of supervision for the model.
\end{abstract}

\section{Introduction}%
\label{s:intro}

Correspondences play an important role in computer vision, with applications to pose estimation, 3D reconstruction, retrieval, image and video editing and many more.
In this paper, we consider the problem of learning dense keypoints for any given type of objects without manual supervision.
Keypoints identify common object parts, putting them in correspondence, and providing a key abstraction in the analysis of the objects' geometry and pose.
While keypoints are usually small in number, dense keypoints~\cite{guler2018densepose} are a generalization that considers a continuous family of keypoints indexed by the surface of a 3D template of the object.
Dense keypoints provide more nuanced information than sparse ones and have found numerous applications in computer vision and computer graphics.

Despite their utility, learning keypoints, especially dense ones, remains labour-intensive due to the need to collect suitable manual annotations.
Because of this, most keypoint detectors are limited to specific object classes of importance in applications, such as humans~\cite{guler2018densepose, rempe2021humor, zhang2021pymaf, kreiss2019pifpaf}.
Methods that generalize to more categories either have limited performance~\cite{kulkarni2019canonical, kulkarni2020articulation}, or require a significant amount of manual annotations for each class~\cite{neverova2020continuous, neverova2021discovering}.
They cannot scale to learning (dense) keypoints for the vast majority of object types in existence.

In contrast, foundation models such as DINO~\cite{caron2021emerging}, CLIP~\cite{radford2021learning}, GPT-4~\cite{chatgpt}, DALL-E~\cite{ramesh2022dalle2}, and Stable Diffusion~\cite{rombach2021stable_diffusion} are trained from billions of Internet images and videos with almost no constraints on the type of content observed.
While these models do not provide explicit information about the geometry of objects, we hypothesise that they may do so \emph{implicitly} and may thus be harnessed to generalize geometric understanding to more object types.

In this paper, we test this hypothesis by utilizing off-the-shelf foundation models to learn automatically high-quality dense keypoints.
Given a single template mesh for an object class (\eg, a horse or a T-Rex) to define the index set for the keypoints, and as few as 1,000 masked example images of the given class, we learn a high-quality image-to-template mapping.

Our method builds on recent advances in self-supervised image-to-image matching algorithms which, by using features from DINO~\cite{caron2021emerging} and the Stable Diffusion encoder~\cite{rombach2021stable_diffusion}, can generalize surprisingly well across images of different modalities or styles, such as natural images, animations or abstract paintings.
Our idea is to reduce the problem of matching images to the 3D template to the one of matching images to \emph{rendered views of the template}.
Namely, we render a view of the 3D template and, given a query location in the source image, we find the corresponding vertex as a visual match on the rendered images.
The template renders are \emph{not} photorealistic, so the matching process emulates the process of manually annotating dense keypoints in prior works~\cite{guler2018densepose,neverova2020continuous}.
We contribute several ideas to robustly pool information collected from different renders of the template, including accounting for visibility.

The approach we have described so far is training-free, as it uses only off-the-shelf components, but it is slow and the resulting correspondences lack spatial smoothness as they are established greedily.
Our second step is thus to use these initial correspondences to supervise a more traditional dense keypoint detector in the form of a canonical surface map~\cite{thewlis17dense,guler2018densepose,kulkarni2019canonical}.
We utilize the Canonical Surface Embedding (CSE) representation of~\cite{neverova2021discovering}, which was designed to learn a mapping for several proximal object classes together (\eg, cow, dog and horse), and can also efficiently represent image-to-template and image-to-image mappings by learning cross-modal embeddings.
The most important result is that we can \emph{outperform} the original manually-supervised model of~\cite{neverova2021discovering} on their animal classes \emph{without} using any supervision.
This means that we can also learn maps for entirely new classes, such as T-Rex or Appa (a flying bison from a TV show), essentially at no cost (\cref{fig:teaser}).

Finally, we note a further use of foundation models for our application: the generation of photorealistic synthetic images of the object.
In particular, we show that a version of Stable Diffusion conditioned on depth can be used to texture the 3D images of the template, significantly narrowing the synthetic-to-real gap.
These images are good enough to be used to supervise the dense pose map \emph{directly}, with full synthetic supervision.
We show that, while this is no substitute for utilizing real images as described above, it does improve the final performance further.

\section{Related work}%
\label{s:related}

\paragraph{Unsupervised image-to-image correspondences.}

Many authors have sought to establish correspondences between images without manual supervision to address the cost of obtaining labels for this task.
Early methods generate training data by applying synthetic warps to images~\cite{arbicon-net, rocco2017convolutional, truong2020glu, melekhov2019dgc, truong2020gocor}, or use cycle consistency losses~\cite{jeon2018parn, truong2021warp, truong2022probabilistic, shtedritski23learning}.
GANs have also been used to supervise dense visual alignment~\cite{peebles2022gangealing}.
Recent advances in self-supervised representation learning~\cite{caron2021emerging, oquab2023dinov2} and generative modelling~\cite{rombach2021stable_diffusion} have boosted the quality of unsupervised semantic correspondences significantly.
For instance, {}\cite{amir21deep} establish correspondences by seeking matches between DINO features, and {}\cite{luo2023dhf, li2023sd4match, hedlin2023unsupervised, tang2023emergent, zhang2023sd-dino} use Stable Diffusion instead.
Similarly,~\cite{Dutt_2024_CVPR, morreale2024neural} use diffusion features to find mesh-to-mesh correspondences.
{}\cite{zhang2023sd-dino} show that DINO and Stable Diffusion features are complementary, the first capturing precise but sparse correspondences and the second the general layout, and propose to combine them.
In our work, we use the SD-DINO~\cite{zhang2023sd-dino} features for matching images.

\paragraph{Animal pose estimation.}

While most works on pose estimation focus on humans~\cite{bourdev09poselets, felzenszwalb08a-discriminatively, newell16stacked, cao17realtime, wei16convolutional, guler2018densepose}, several authors have attempted to estimate the pose of animals by detecting~\cite{zhang14part-based}, matching~\cite{kanazawa16warpnet:} or reconstructing~\cite{wu23magicpony, wu21dove:} them, or predicting the parameters of parametric models~\cite{zuffi173d-menagerie:,zuffi18lions,zuffi2019three}.
However, these methods do not scale well as they need annotations for each type of animal considered.
Our method is most similar to~\cite{kulkarni2020articulation, kulkarni2019canonical} in that we only require a template shape and a collection of images to learn image-to-shape correspondences.
However, our method achieves much better performance, while still using fewer images for training.

\paragraph{Image-to-template correspondences.}

Finding correspondences between images and a 3D template is useful for understanding the geometry of deformable objects, with several applications.
For instance, it is used in biology to study the behaviour of animals~\cite{poseofflies, waldmann20233d}.
Most prior works focus on humans~\cite{guler2018densepose, rempe2021humor, zhang2021pymaf, kreiss2019pifpaf} due to the availability of large-scale datasets of densely annotated image-template pairs, such as DensePose-COCO~\cite{guler2018densepose}.
Similar datasets exist for animals, such as DensePose-LVIS~\cite{neverova2021discovering}, but are much smaller and still only cover a handful of animal classes.
To learn image-to-shape correspondences, {}\cite{kulkarni2020articulation, kulkarni2019canonical} parametrise a 3D shape as a 2D $uv$ map, and use cycle consistency to try to learn correspondences automatically, whereas~\cite{kulkarni2020articulation} also learn to predict articulation.
Similarly to~\cite{neverova2021discovering}, we use the CSE representation for learning the correspondences.
However, differently from~\cite{neverova2021discovering}, our method does not rely on any human-annotated data.

\section{Method}%
\label{s:method}

\begin{figure}[t]
\begin{center}
\includegraphics[width=1.0\textwidth]{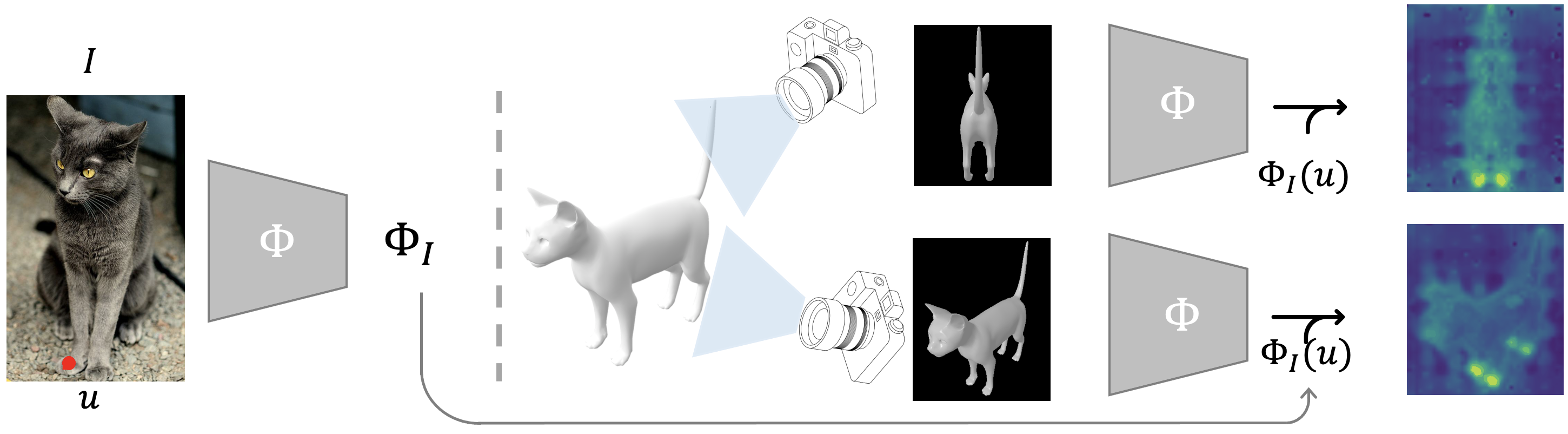}\caption{\textbf{Image-to-template correspondences using 2D renderings.}
Using an unsupervised semantic correspondence method, we can find correspondences between an image of an object and a rendering of its 3D template.
Here we show the similarity heatmap from the source location (annotated in red) to all pixel locations in the target image using SD-DINO~\cite{zhang2023sd-dino}.}\label{fig:similarity}
\end{center}
\end{figure}

In this section, we describe \method, our method for learning dense keypoints without manual supervision.
First, in \cref{sec:dense-keypoints} we recall the notion of dense keypoints, canonical surfaces and canonical surface maps.
Then, in \cref{sec:unsup-matching} we discuss using self-supervised features to establish dense semantic correspondences between pairs of images, lift those to dense keypoints in \cref{sec:image-to-3d-matching}, and use the latter to supervise a canonical map in \cref{sec:densepose-method}.
Finally, in \cref{sec:realism}, we show how an image generator can produce realistic views of the template, which further improves results.

\subsection{Canonical surface maps}%
\label{sec:dense-keypoints}

Let $I \in \mathbb{R}^{3 \times \Omega}$ be an image supported by the grid
$
\Omega =\{1,\dots,H\} \times \{1,\dots,W\}.
$
The image contains an object of a given type, such as a cat, and the goal is to assign an identity to each pixel $u \in U_I$ of the object, where $U_I \subset \Omega$ is the object mask in image $I$.
The identification is carried out by a mapping $f_I : U_I \rightarrow M$ that assigns each pixel $u$ to a corresponding index $f_I(u)$ in a set $M$.
The set $M \subset \mathbb{R}^2$ is a 2D surface embedded in $\mathbb{R}^3$, and is interpreted as a (fixed and rigid) 3D template of the object.
The template $M$ is also called a \emph{canonical surface} and the function $f_I$ a \emph{canonical surface map}.
The same canonical surface $M$ is shared by all objects of that category.
In this way, by mapping two images $I$ and $J$ to the same template, one can also infer a mapping between the images.

In practice, we approximate the surface $M$ by a mesh supported by a finite set of $K$ vertices
$
V = \{x_1,\dots,x_K\} \subset M
$
and triangular faces $F$.
Hence the canonical map is a function
$
f_I : \Omega \rightarrow V \subset M
$.
This slightly simplifies the formulation as both index sets $\Omega$ and $V$ are finite.
We also note that the value $f_I(u)$ is undefined if pixel $u \in \Omega - U_I$ does not belong to the object.

In prior works, learning the canonical map $f$ often requires hundreds of thousands of manually specified image-to-template correspondences.
In the next sections, we will show how to learn this mapping \emph{automatically} instead.

\subsection{Unsupervised image-to-image correspondences}%
\label{sec:unsup-matching}

In order to learn the canonical map $f$ automatically, we start by establishing correspondences between pairs of images $I$ and $J$ in an unsupervised fashion.
We do so by first computing $D$-dimensional dense features $\Phi \in \mathbb{R}^{D \times \Omega}$ using a pre-trained network.
Then, we associate each query location $u$ in the source image $I$ to the location $v_u$ in the target image $J$ with the most similar feature vector based on the cosine similarity, \ie,
$$
v_u = \argmax_{v\in\Omega} S_{IJ}(u, v)
~~~\text{where}~~~
S_{IJ}(u,v)
=
\dfrac
{\Phi_I(u) \cdot \Phi_J(v)}
{\left \| \Phi_I(u) \right \|_{2} \left \| \Phi_J(v) \right \|_{2}}.
$$
The quality of the correspondences depends on the quality of the feature extractor $\Phi$.
In particular, by using the unsupervised features by~\cite{zhang2023sd-dino}, it is possible to establish good correspondences between a (real) image $I$ of the object and a \emph{rendering} of the 3D template $M$.

This is illustrated in \cref{fig:similarity}, where we show the cosine similarity heatmaps between a feature at a query location $u$ of in the source image $I$ and all locations in several 3D renders of the template.
While the correspondences correctly identify the type of body part (paw), two problems are apparent:
(i) there is left-right ambiguity, which is common for unsupervised semantic correspondence methods~\cite{zhang2023telling},
(ii) when the correct match is not visible (as on the top of \cref{fig:similarity}, where only the back paws are visible), the correspondence will always be wrong.
In the next section, we lift these image-based correspondences into correspondences with the template $M$, which also alleviates these issues.

\subsection{Unsupervised image-to-template correspondences}%
\label{sec:image-to-3d-matching}

\begin{figure}[t]
\begin{center}
\includegraphics[width=1.0\textwidth]{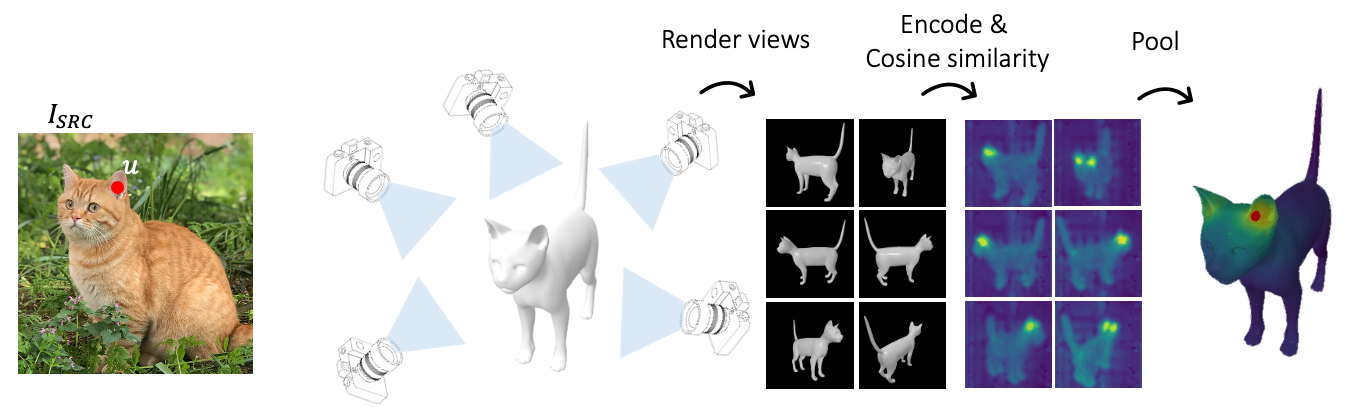}
\caption{\textbf{Zero-shot image-to-template correspondences.}
From left to right:
an image $I$ with a selected pixel $u$;
several views $J_i$ of the synthetic template;
corresponding  renderings and similarities $S_{IJ_i}(u,v)$ as functions of the target locations $v \in \Omega$;
the final similarities $\Sigma_I(u)$  visualized as a heatmap on top of the canonical surface $M$.
The maximizer of the latter (red dot) identifies the vertex $x_k$ that best corresponds to the selected pixel $u$ in the source image $I$ (\ie, base of the left ear of the cat).
}%
\label{fig:pesudo-gt}
\end{center}
\end{figure}

Given the source image $I$ and a pixel $u$, we now consider the problem of finding the vertex $x_k$ in the template $M$ that best represents it.
In order to do so, we develop a similarity measure between pixels and vertices, utilizing the image-to-image similarity metric of \cref{sec:unsup-matching}.
We first generate $N$ different views
$
J_i = \operatorname{Rend}(M, c_i)
$,
$i=1,\dots,N$,
of the canonical surface $M$ by rendering it from viewpoints $c_i$ (camera parameters).
For each view $J_i$, we project each vertex $x_k$ to its closest location in the mask $U_{\!J_i}$, defining
\begin{equation}\label{eq:reprojection}
v_i(k) = \argmin_{v\in U_{\!J_i}} \|v - \pi(x_k, c_i) \|,
\end{equation}
where $\pi(x_k, c_i)$ is the camera projection function.
We also denote by $V_i \subset V$ the subset of vertices $x_k$ that are visible in view $J_i$.

Given this notation, we can define a new score $\Sigma$ measuring the compatibility between each location $u$ in the source image $I$ and each vertex $x_k \in V$ in the canonical surface, and corresponding matches $\tilde x$, as follows:
\begin{equation}\label{eq:pseudo-gt}
\Sigma_I(u, x_k)
= \operatornamewithlimits{pool}_{i : x_k \in V_i} S_{IJ_i}(u, v_i(k)),
~~~
\tilde x(u, I) = \argmax_{x_k \in V} \Sigma_I(u, x_k).
\end{equation}
The goal of the pooling operator is to assess the compatibility between pixel $u$ and vertex $x_k$ into a single score that consolidates the information collected from the different viewpoints $c_i$.
Note that only the views where the vertex is visible are pooled.
In practice, we set the pooling operator to average or max pooling.

\paragraph{Illustration.}

\Cref{fig:pesudo-gt} illustrates the similarity maps $S_{IJ_i}$ between the source image $I$ and various views $J_i$ of the rendered 3D object, as well as the result $\Sigma_I$ of mapping and pooling them on the canonical surface itself.
We see that the correct semantic parts on the shape are identified (ears), and the base of the left ear is selected as the most similar to the query $u$.

\begin{wrapfigure}[7]{r}{0.4\textwidth}
\includegraphics[width=0.4\textwidth]{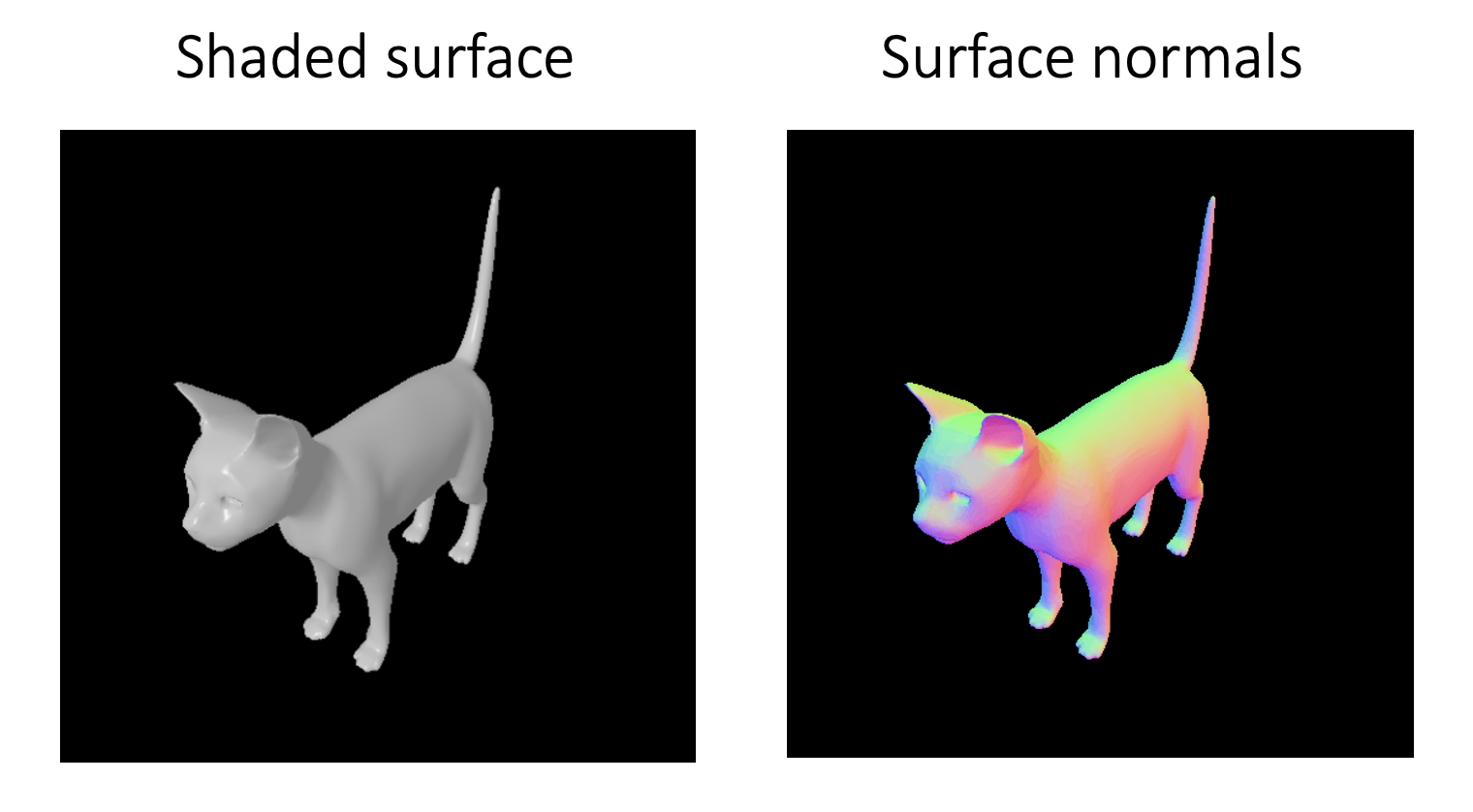}
\end{wrapfigure}

\paragraph{Template realism and rendering function.}

The template $M$ captures the typical shape of the object.
Mathematically, its main purpose is to define the \emph{topology} of the object's surface, but the latter is usually topologically equivalent to a sphere~\cite{thewlis17dense}.
In dense pose~\cite{neverova2020continuous} there are two reasons for not using a sphere.
The first is that the \emph{metric} of the template surface can be used to regularize correspondences (e.g., by capturing an approximate notion of how far apart physical points are).
The second is that \emph{renders} of the template are given to human annotators to establish correspondence with the template.
Our method can be seen as automatizing the annotation step.
Just like for manual annotation, it does \emph{not} require a photorealistic rendition of the template.
However, it does not mean that \emph{all renditions are equally good} from the viewpoint of the matching network.
Inspired by previous work on image generation~\cite{chen23fantasia3d_mine}, we find that rendering a \emph{normal map} of the 3D template results in better matches than rendering a shaded version of the same (see the embedded figure).
We discuss realistic rendering in \cref{sec:realism}.

\subsection{Unsupervised canonical surface maps}%
\label{sec:densepose-method}

\begin{figure}[t]
\begin{center}
\includegraphics[width=0.95\textwidth]{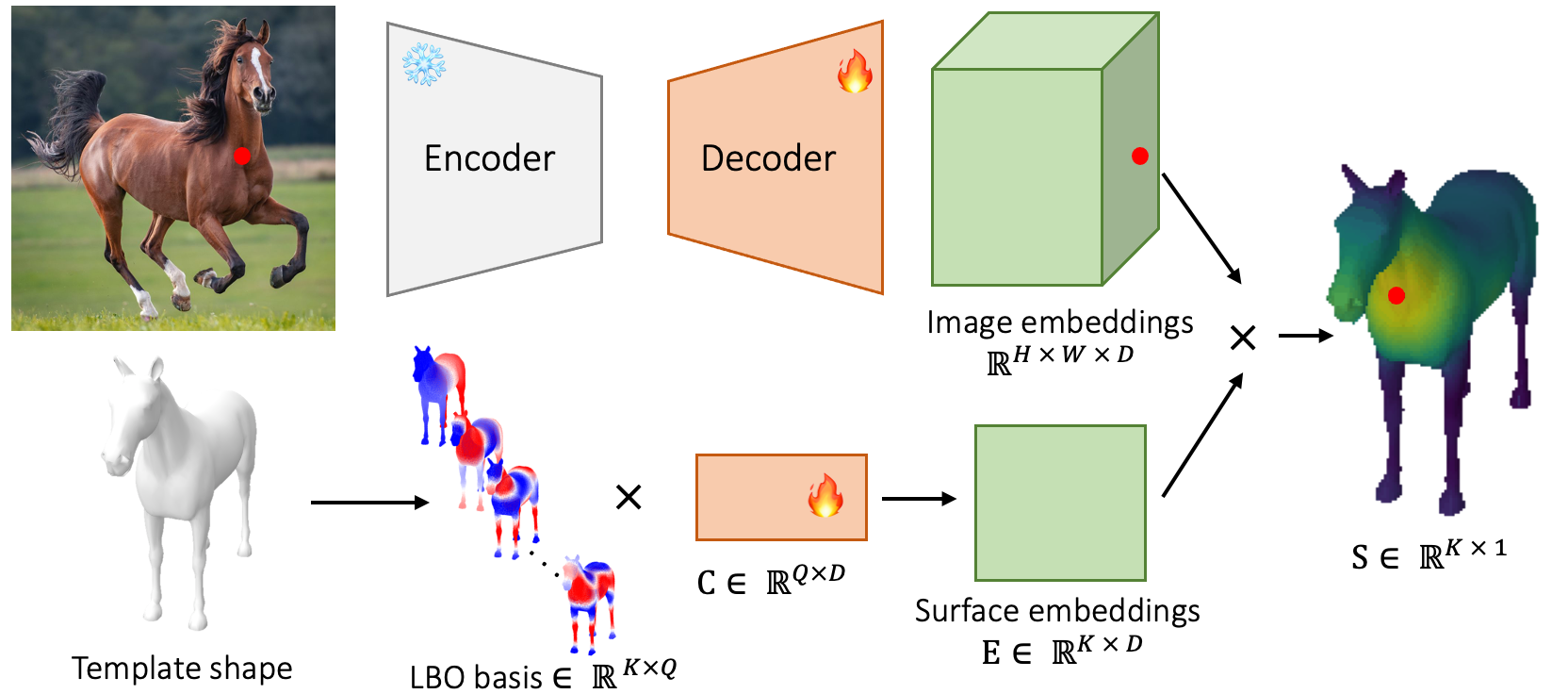}
\caption{\textbf{CSE dense pose predictor.} We jointly train a deep network $\Phi$ and a matrix $C$, that transforms LBO eigenvectors to a shared $D$-dimensional space. We use pseudo-ground truth, obtained as described in~\cref{sec:image-to-3d-matching} for supervision. The image encoder is a frozen pre-trained DINO ViT, and the decoder we learn is a CNN.}\label{fig:method}
\end{center}
\end{figure}

Here, we show how to learn the  canonical surface map $f$ from the image-to-template correspondences constructed in \cref{sec:image-to-3d-matching}.
An overview is in \cref{fig:method}.

\paragraph{Continuous Surface Embeddings.}

Following~\cite{neverova2020continuous, neverova2021discovering}, we represent the map $f$ via Continuous Surface Embeddings (CSEs).
CSE assign embedding vectors $e_I(u), e(x_k) \in \mathbb{R}^D$ to each image pixel $u$ and each mesh vertex $x_k$ so that the correspondences are defined by maximizing their similarity:
\begin{equation}\label{eq:csm}
f_I(u) = \argmax_{x_k \in V} p(x_k | u, I),
 ~~\text{where}~~
p(x_k | u, I)
=
\dfrac
{\exp(\langle e_I(u), e(x_k) \rangle)}
{\sum^{K}_{t=1}\exp(\langle e_I(u), e(x_t) \rangle)}.
\end{equation}
Learning the CSE model thus amounts to learning the vertex embeddings $e(x_k)$ as well as a corresponding dense feature extractor $e_I$.

The vertex embeddings are optimized directly as there is a single template mesh.
However, due to the large number of vertices, they are not assumed to be independent but to form a smooth (vector) function over the mesh surface.
This way, the number of parameters required to express them can be reduced significantly.
Collectively, all embeddings $e(x_k)$, $k=1,\dots,K$, form an embedding matrix $E \in \mathbb{R}^{K \times D}$.
The latter is decomposed as the product $E = UC$ where $U \in \mathbb{R}^{K \times Q}$ is a smooth and compact functional basis (akin to Fourier components defined on the mesh) such that $Q \ll K$.
Following~\cite{neverova2020continuous, neverova2021discovering}, we use the lowest eigenvectors of the Laplace-Beltrami operator (LBO) of the mesh $M$ to form $U$.
The only learnable parameters are $C \in \mathbb{R}^{Q\times D}$, which are few.

The other component is the feature extractor $e_I(u)$.
For this, we encode the source image $I \in \mathbb{R}^{3 \times H \times W}$ with a frozen self-supervised encoder (DINO~\cite{oquab2023dinov2}), before decoding it with a CNN back to the original resolution to the required feature tensor $e_I \in \mathbb{R}^{D \times H \times W}$.

\paragraph{Training formulation.}

We train our model with several losses.
The first one simply uses the pseudo-ground-truth correspondences $\tilde x(u,I)$ of \cref{eq:pseudo-gt} in \cref{eq:csm}.
We  pose this as a classification problem, where given a query location $u$ in image $I$ is matched probabilistically to the pseudo-ground-truth $\tilde x(u,I)$ using the cross-entropy loss:
$
\mathcal{L}_\text{pseudo}(I)
=
- \frac{1}{|U_I|}\sum_{u \in U_I} \log p(\tilde x(u, I) | u, I).
$
Additionally, we use the distance-aware loss of~\cite{neverova2021discovering}:
$
\mathcal{L}_\text{dist}(I)
=
- \frac{1}{|U_I|}\sum_{u \in U_I} \sum_{x \in V} d(x, \tilde x)p(x| u, I),
$
where $d(x, \tilde x)$ is the geodesic distance between the vertex $x$ and the pseudo ground-truth $\tilde x$, which discourages placing probability mass far from $\tilde x$.

As noted in \cref{sec:image-to-3d-matching}, there is some ambiguity because of the symmetry of most animals, where the matches are confused between left and right.
To reduce this ambiguity, we use a cycle consistency loss~\cite{neverova2020continuous}, where given a starting location $u$, we match it to a vertex $x_k$ in the mesh, and then matching that back to the image, results in the probability
$
p(v|u,I) = \sum_{x_k \in V}  p(v| x_k, I) p(x_k | u, I)
$
of landing to a location $v$.
Here $p(v |x_k, I)$ is the same as $p(x_k|v,I)$ from \cref{eq:csm} up to renormalization.
We close the image-shape-image cycle and supervise using
$
\mathcal{L}_\text{cyc}(I)
=
\sum_{u \in U_I} \sum_{v \in U_I} \|u -  v\| p(v|u).
$

To further reduce the left-right ambiguity, we assume that the template $V$ has a bilateral symmetry (true for most categories).
Then, for each vertex $x \in V$, let $x_F \in V$ be its symmetric one (for meshes which are not exactly symmetric, we let $x_F$ be the closest approximation to the symmetric version of $x$).
Given an image $I$ and a pixel $u$, denote by $I_F$ and $u_F$ their horizontal flips.
Suppose that $u$ is the pixel that corresponds to vertex $x$ in image $I$.
Then one can show~\cite{thewlis18modelling} that pixel $u_F$ must correspond to vertex $x_F$ in image $I_F$, leading to the loss:
$
\mathcal{L}_\text{eq}(I) 
=
\frac{1}{|U_I|}
\sum_{u \in U_I}
\sum_{x \in V}
|p(x|u, I) - p(x_F|u_F, I_F)|.
$

\subsection{Increasing the realism of synthetic data}%
\label{sec:realism}

\begin{figure}[t]
\begin{center}
\includegraphics[width=1.0\textwidth]{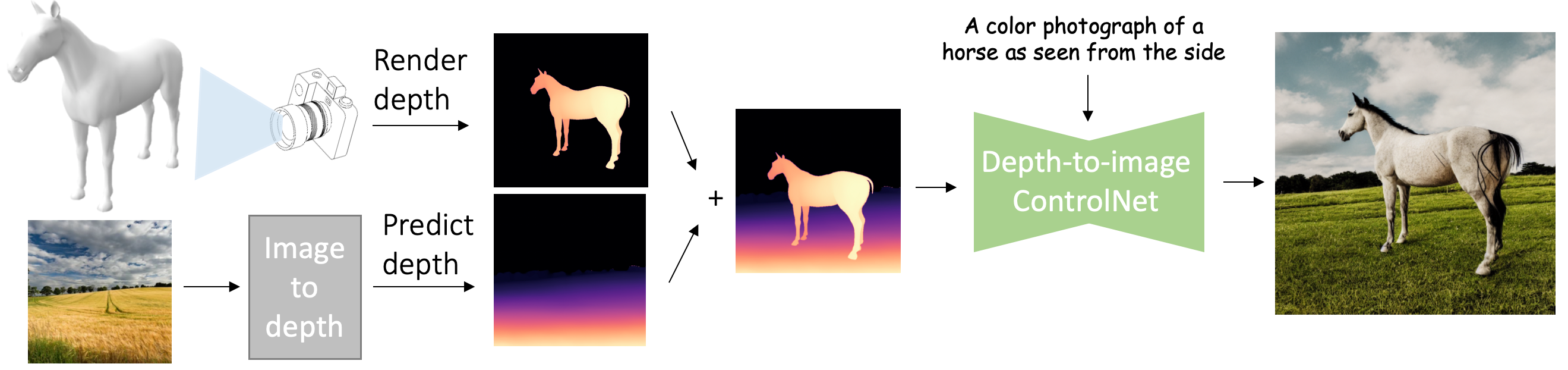}
\caption{\textbf{Realistic rendering of the template.} We create synthetic data for pixel-vertex correspondences by generating photorealistic images from depth renders. The corresponding vertices we obtain from the projections of vertices on the image.}\label{fig:synth-generation}
\end{center}
\end{figure}

The renders $J_i$ of the 3D template $M$ can be used to supervise the canonical map $f$ directly because we \emph{know} the 2D location $v_i(k)$ of each vertex $x_k$ in image $J_i$ based on \cref{eq:reprojection}.
With this, we can write the loss:
\begin{equation}\label{eq:direct}
\mathcal{L}_\text{syn}(J_i, v_i)
= - \frac{1}{K} \sum_{k=1}^K\sum_{i : V_k \in i}  \log p(x_k | v_i(k), J_i)
\end{equation}
The very limited diversity and realism of the renders $J_i$ makes this loss uninteresting, but, as shown in \cref{fig:synth-generation}, we can  use a powerful image generator to significantly augment the realism of such renders.

To do this, we first render a depth image of the template $M$ from a random viewpoint $c$.
We also sample a random background image and predict its depth using~\cite{depthanything}, blend the foreground and background depth images, and use the depth-to-image ControlNet~\cite{zhang2023adding} of~\cite{depthanything} to generate photorealistic image $J_i$ of the template.
We prompt the depth-to-image model
(i) using the object's class name, e.g., ``horse'', and 
(ii) specifying the viewpoint (``front'', ``side'', or ``back''), which we heuristically obtain from the camera location w.r.t.~the 3D template $M$.

The results are photo-realistic renders $J_i$ of the template.
Note that the appearance of different renders is not consistent, but this is a feature rather than an issue in our case because we need to learn an image-to-template map, which is invariant to details of the appearance.
The main limitation is that the template is fixed, so there is no diversity in terms of pose and 3D shape.
Hence, we expect loss \cref{eq:direct} to be complementary rather than substitutive of the one above.

\subsection{Learning formulation}%
\label{s:learning-formulation}

Given a dataset $\mathcal{D}$ of masked training images of the object, our loss is:
$$
\mathcal{L}
=
\frac{1}{|\mathcal{D}|}
\sum_{I \in \mathcal{D}}
\left(
\alpha\mathcal{L}_\text{pseudo}(I) +
\beta\mathcal{L}_\text{cyc}(I) +
\gamma\mathcal{L}_\text{dist}(I) +
\delta\mathcal{L}_\text{eq}(I)
\right)
+
\zeta
\sum_{i=1}^N \mathcal{L}_\text{syn}(J_i, v_i),
$$
where $\alpha,\beta,\gamma,\delta$ and $\zeta$ are coefficients set empirically.
\section{Experiments}%
\label{sec:experiments}

\definecolor{darkgreen}{rgb}{0,0.5,0}
\newcommand{\red}[1]{\textcolor{red}{#1}}
\newcommand{\green}[1]{\textcolor{darkgreen}{#1}}
\begin{table}[t]
\centering
\resizebox{0.99\textwidth}{!}{%
\begin{tabular}{lcccccccc|c}
\toprule
Method & Supervision & Horse & Sheep & Bear & Zebra & Cow & Elephant & Giraffe & \textbf{Average}\\
\midrule
CSE~\cite{neverova2021discovering} & S & 24.1 & 32.0 & 35.7 & 24.9 & 25.4 & 26.1 & \textbf{18.0} & 26.6\\
Zero-shot${}_\mathrm{SD-DINO}$ & U & 37.2 & 41.4 & 48.2 & 32.0 & 32.3 & 36.0 & 26.3 & 36.2\\
\method (Ours) & U & \textbf{23.3} & \textbf{30.3 }& \textbf{33.0} & \textbf{23.3} & \textbf{22.7} & \textbf{23.9} & 18.1 & \textbf{24.9} \\
\bottomrule
\end{tabular}%
}
\caption{\textbf{Evaluation on DensePose-LVIS.}
We compare the supervised (S) method of~\cite{neverova2021discovering} to our unsupervised method (U) and our adaptation of SD-DINO to DensePose described in~\cref{sec:image-to-3d-matching}. We evaluate~\cite{neverova2021discovering} using their published weights.
We measure geodesic error (\textit{lower is better}).}%
\label{tab:densepose-eval}
\end{table}
\begin{table}[t]
\centering
\resizebox{0.65\textwidth}{!}{%
\begin{tabular}{lcccc|c}
\toprule
Method & Supervision & Cow & Sheep & Horse & \textbf{Average}\\
\midrule
Rigid-CSM~\cite{kulkarni2019canonical} & S & 28.5 & 31.5 & 42.1 & 34.0\\
A-CSM~\cite{kulkarni2020articulation} & S & 29.2 & 39.0 & 44.6 & 37.6\\
CSE~\cite{neverova2021discovering} & S & 51.5 & 46.3 & 59.2 & 52.3\\
\midrule
Rigid-CSM~\cite{kulkarni2019canonical} & U & 26.3 & 24.7 & 31.2 & 27.4 \\
A-CSM~\cite{kulkarni2020articulation} & U & 26.3 & 28.6 & 32.9 & 29.3\\
\method (Ours)$_\mathrm{im2im}$ & U & 69.1& 55.9  & 58.7& 61.2\\
\method (Ours)$_\mathrm{im2m2im}$ & U & \textbf{73.5} & \textbf{73.5}& \textbf{63.1}& \textbf{70.0}\\
\bottomrule
\end{tabular}%
}
\caption{\textbf{PCK-Transfer on PF-Pascal.}
We compare against prior work on image-to-image semantic correspondences. We predict image-to-image correspondences either by directly predicting the correspondences or by performing image-to-vertex-to-image matching. We use the reported numbers from~\cite{neverova2021discovering, kulkarni2020articulation, kulkarni2019canonical}, and evaluate using PCK-0.1.
}\label{tab:pf-pascal-eval}
\end{table}
\begin{figure}[t]
\begin{center}
\includegraphics[width=0.98\textwidth]{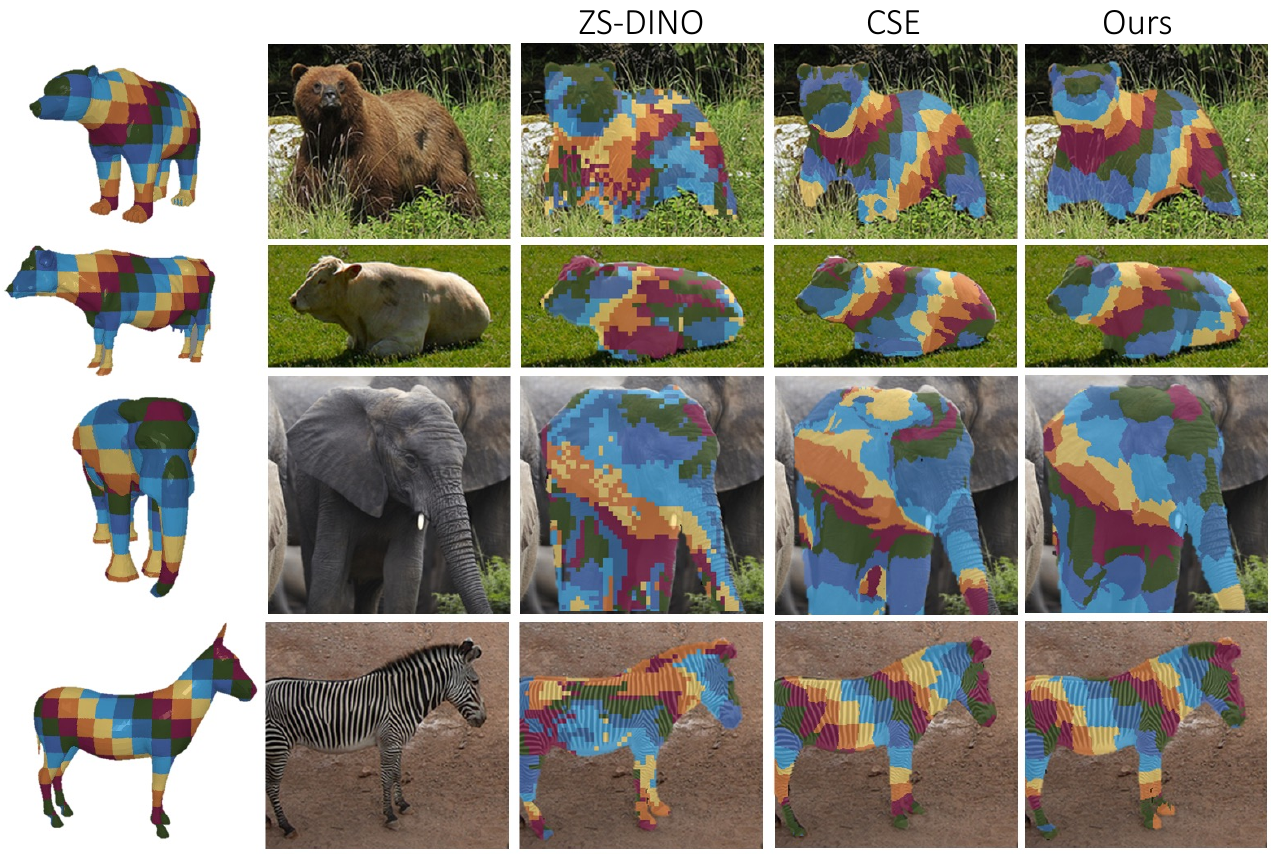}
\caption{\textbf{Mapping a textured mesh.} We map a textured mesh over the image using the predicted dense correspondences.}\label{fig:texture}
\end{center}
\end{figure}
\begin{table}[t]
\centering
\begin{minipage}[t]{0.52\linewidth} %
\centering
\resizebox{0.9\textwidth}{!}{%
\begin{tabular}{llc}
\toprule
Ablation & & DensePose-LVIS \\
\midrule
& Ours & \textbf{24.9}\\
\midrule
Pooling & \texttt{mean} (instead of \texttt{max})& 26.6 \\
\midrule
\multirow{3}{*}{Data} &
w/o Synth data & 25.6\\
& w/o LVIS & 33.8 \\
& w/o pseudo GT & 65.8 \\
\bottomrule
\end{tabular}}
\caption{\textbf{Data and other ablations.}
First, we ablate the pooling function used to construct $\Sigma$, and evaluate \texttt{mean} instead of \texttt{max} pooling.
Next, we compare the data we use --- we remove (1) synthetic data, (2) natural images,  (3) the pseudo ground truth for natural images.
In (3), we still use natural images for $\mathcal{L}_\text{eq} \text{ and } \mathcal{L}_\text{cyc}$, but do not use the pseudo-GT for $\mathcal{L}_\text{pseudo} \text{ and } \mathcal{L}_\text{dist}$.}%
\label{tab:results-ablations-data}
\end{minipage}
\hfill
\begin{minipage}[t]{0.45\linewidth} %
\centering
\resizebox{\textwidth}{!}{%
\begin{tabular}{llc}
\toprule
Ablation & & DensePose-LVIS \\
\midrule
& Ours & \textbf{24.9}\\
\midrule
\multirow{5}{*}{Losses} &  w/o $\mathcal{L}_\text{pseudo}$ &  27.3 \\
&  w/o $\mathcal{L}_\text{dist}$ &  25.8 \\
& w/o $\mathcal{L}_\text{eq}$  & 25.6 \\
&  w/o $\mathcal{L}_\text{cyc}$ & 25.8  \\
& w/o $\mathcal{L}_\text{eq}$ \& $\mathcal{L}_\text{cyc}$ & 26.1 \\
\bottomrule
\end{tabular}}
\caption{\textbf{Ablating the losses.}
We assess the contributions of the losses removing them one at a time.
On the last row, we remove $\mathcal{L}_\text{eq} \text{ and } \mathcal{L}_\text{cyc}$ to show that although both address symmetry, they are complementary.
Ablating $\mathcal{L}_\text{syn}$ falls under \cref{tab:results-ablations-data} (w/o Synth data) as it uses different data.
We show the average score over all classes.}%
\label{tab:results-ablations-losses}
\end{minipage}
\end{table}

\begin{figure}[t]
\begin{center}
\includegraphics[width=1.0\textwidth]{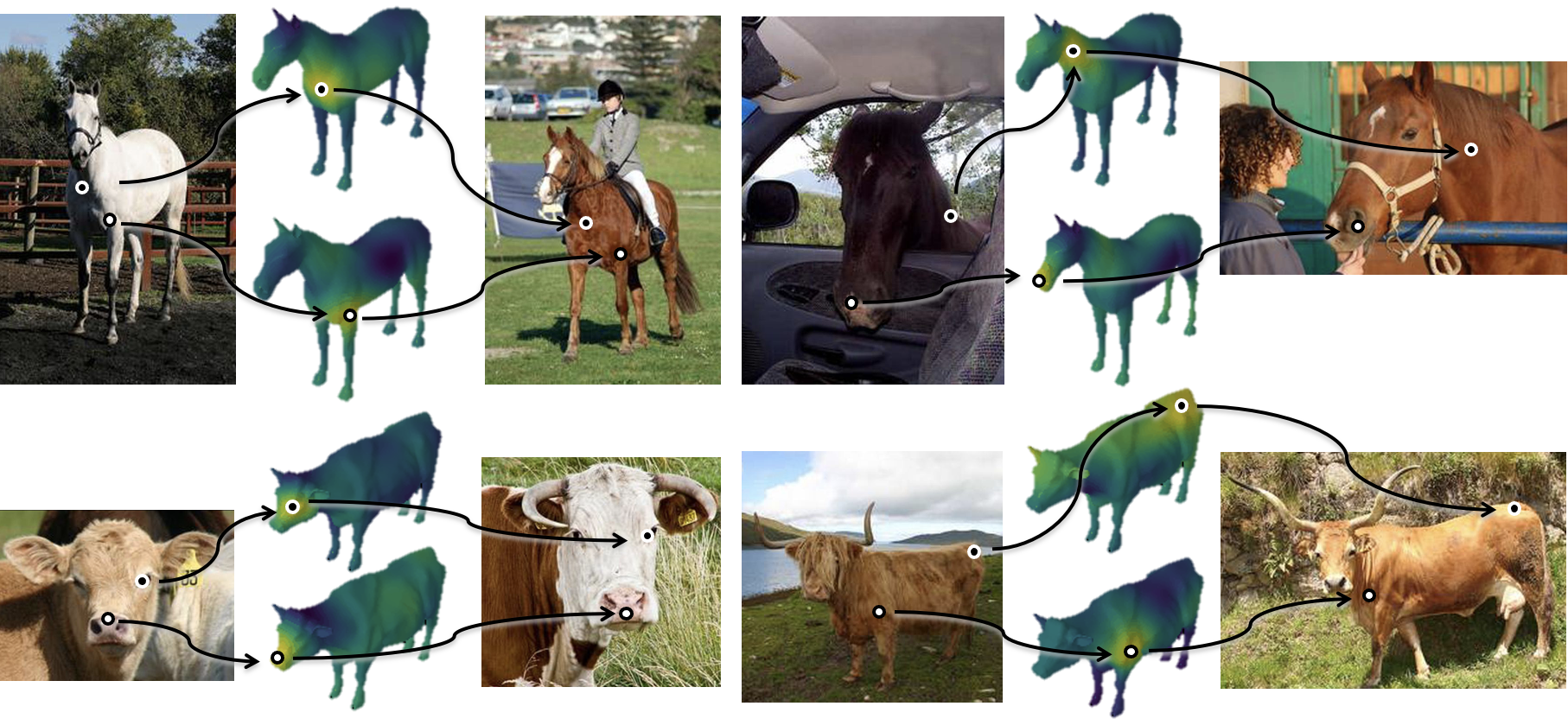}
\caption{\textbf{Image-to-image correspondences.} We show image-to-image correspondences on PF-PASCAL, which we find using pixel-to-vertex-to-pixel matching. The heatmaps on the shape show the similarity from the source image location to every vertex.}\label{fig:im2im}
\end{center}
\end{figure}

We evaluate our method for learning canonical maps automatically, without manual keypoints supervision, against unsupervised and supervised prior work.

\subsection{Implementation details}%
\label{sec:details}

To obtain the image-to-shape similarities $\Sigma$, we use $N=72$ renderings of the template shape (using the surface normal style), render the surface normals, and compute the image-to-image similarities $S$ using the features from~\cite{zhang2023sd-dino}.
For each rendered image, we automatically get the pixel-to-vertex matches from the camera projection function.
Finally, we aggregate the similarities across all views using \texttt{max} pooling.
During training, we randomly select 100 foreground points from each image and their corresponding vertices from $\Sigma$.
Following~\cite{neverova2021discovering}, we use the lowest $Q = 64$ eigenvectors of the LBO of the template mesh and use dimensionality $D=16$ for the joint image-shape embedding space.
For the image encoder $\Phi$, we use a frozen DINO-v2~\cite{oquab2023dinov2} backbone, followed by a decoder consisting of 5 convolutional layers.
We did not do any tuning of the parameters in the final formulation of the loss.
We used values for the loss hyper-parameters that make the losses roughly of similar scale, namely $\alpha=0.1, \beta=0.002, \gamma=0.002, \delta=0.001, \zeta=0.1$.
All models and code will be released upon acceptance of the paper.

\subsection{Training data}

For most of our evaluations, we consider the DensePose-LVIS dataset~\cite{neverova2021discovering}, which applies DensePose to a variety of animal classes.
Of those, we consider the horse, sheep, bear, zebra, cow, elephant and giraffe classes\footnote{For the classes cat and dog, we could not obtain the 3D templates from~\cite{neverova2021discovering} and could therefore not use the annotated image-to-template correspondences for evaluation.}.
The DensePose-LVIS data contains a total of 6k images of these categories, as well as a reference 3D template for each category.
Knowledge of the 3D template is necessary to interpret the annotations in the dataset, as well as to compute the geodesic distances required for evaluation.
Every animal has up to three manually annotated pixel-template correspondences to the corresponding template.
We use these annotations only for evaluation.
We only train our models on cropped animals from DensePose-LVIS~\cite{neverova2021discovering}.
The number of instances for each class varies from 2,899 for horses (most) to 735 for bears (least).
In comparison, competing methods train on more images.
{}\cite{kulkarni2020articulation} train on combined PASCAL and ImageNet images, and the supervised method of~\cite{neverova2021discovering} is trained on DensePose-LVIS and DensePose-COCO, the latter of which consists of 5 million image-to-template annotations for humans.
Likely due to the density of the pseudo-ground truth, our method only needs a much smaller amount of data and can be trained with as few as a few hundred images (\eg, bear class from DensePose-LVIS, or the models we show in~\cref{fig:teaser}).
Similarly to CSM~\cite{kulkarni2019canonical, kulkarni2020articulation} and CSE~\cite{neverova2021discovering}, we use masks for training.

\subsection{Evaluation}%
\label{sec:data}

We evaluate our models and CSE~\cite{neverova2021discovering} on DensePose-LVIS using the geodesic error.
We normalize the maximum geodesic distance on each mesh to 228, following~\cite{guler18densepose:, neverova2020continuous}, and use a heat solver to obtain all vertex-to-vertex geodesic distances.
Additionally, we use PF-PASCAL~\cite{ham2017proposal} to evaluate our model on keypoint transfer.
PF-PASCAL consists of pairs of images with annotated salient keypoints (\eg, left eye, nose, \etc), and we evaluate image-to-mesh-to-image correspondences using the image-to-image annotations.
We use the test split of~\cite{zhang2023sd-dino} and evaluate using PCK 0.1, following prior work~\cite{zhang2023sd-dino, neverova2021discovering, kulkarni2020articulation}.

\paragraph{Image-to-shape correspondences.}%
\label{sec:results-image-to-shape-matching}

In \cref{tab:densepose-eval} we compare the quality of the image-to-template correspondences established by \method, by our zero-shot method based on SD-DINO, and by CSE~\cite{neverova2021discovering}, which is supervised, on the DensePose-LVIS dataset.
The most important result is that \method learns better canonical maps than CSE despite using no supervision.
While the training images of \method and CSE are the same and the pseudo-ground truth is noisy (Zero-shot$_\text{SD-DINO}$ is what we use for supervision), this result can be explained by the fact that our automated supervision is significantly denser than the manual labels collected by~\cite{neverova2021discovering} (just three per image).

We show some qualitative results on this dataset in \cref{fig:texture}, where we color every point on the image according to the corresponding color on the mesh.
The regularity of the remapped texture illustrates the quality of the correspondences.
Once more, the learned canonical map (ours) is significantly better than the pseudo-ground truth (SD-DINO).
Compared to CSE, \method performs similarly and tends to have a more regular structure on the heads.
We show more qualitative evaluations and failure cases in the Appendix.

\paragraph{Ablation study.}%
\label{sec:ablations}

We ablate the components of our method in \cref{tab:results-ablations-data} and \cref{tab:results-ablations-losses}.
When using mean pooling instead of max pooling for obtaining the pseudo-ground-truths, we see a significant drop in performance.
This is because image-to-image correspondences are more reliable when the objects are in similar poses, and with max pooling we only get contribution from the most similar view.
Next, we look at the importance of different sources of supervision.
When we remove the pseudo-ground-truth from our synthetic pipeline (\cref{sec:realism}), the model loses performance. When we exclude all natural images (w/o LVIS), and only train on synthetically generated data, we see a more pronounced drop in performance.
Finally, we exclude the pseudo-ground-truth from SD-DINO (\cref{sec:image-to-3d-matching}) and thus only train with the synthetic data and the cycle consistency and equivariance losses on natural images.
In this case, the model learns a degenerate solution, where for natural images it only predicts vertices on one side of the shape (\ie, left or right).
Such degenerate solutions have been observed by~\cite{truong2021warp} for unsupervised image-to-image matching when using cycle consistency losses.
This shows the importance of using our pseudo-ground-truth.
Finally, we ablate the losses we use in~\cref{tab:results-ablations-losses}.
All losses are necessary for the final performance.
We find that $\mathcal{L}_\text{pseudo}$, where we frame pixel-to-vertex matching as a multi-class classification problem, has a bigger contribution than the distance-aware loss $\mathcal{L}_\text{dist}$ of~\cite{neverova2021discovering}.
Additionally, we see that both losses that address symmetry, $\mathcal{L}_\text{eq}$ and $\mathcal{L}_\text{cyc}$, improve performance, and are complementary to each other.

\paragraph{Keypoint transfer.}%
\label{sec:results-keypoint-transfer}

Next, we evaluate \method on the PF-PASCAL~\cite{zuffi173d-menagerie:} in \cref{tab:pf-pascal-eval}.
In this dataset, one evaluates the quality of image-to-image correspondences instead of image-to-template.
There are two ways of using our method to induce image-to-image correspondences.
The first is to use the learned canonical maps, transferring points from one image to the template and then back to the other image.
The second is to directly match the image-based CSE embedding learned in \cref{sec:densepose-method}.
The key findings from our results are:
(1) \method outperforms the supervised versions of
Rigid~\cite{kulkarni2019canonical} and Articulated~\cite{kulkarni2020articulation} CSM, as well as CSE~\cite{neverova2021discovering} and greatly outperforms all unsupervised approaches.
(2) The image-to-image correspondences induced by the canonical maps are significantly better than the ones induced by the image-based CSE embeddings, once again illustrating the importance of the canonical maps. We show qualitative examples in~\cref{fig:im2im}.

\paragraph{Novel classes.}

\method can be trained on any class as long as there is a collection of a few hundred images and a suitable template mesh. In \cref{fig:teaser} we show qualitative results from a model we train on two classes --- T-Rex and Appa, a six-legged flying bison from Avatar: The Last Airbender, using 480 and 180 manually collected images, respectively. We extract masks for training using the open vocabulary segmentation method of~\cite{liu2023grounding}. Although the 3D template for Appa is toy-like and does not resemble the images closely, and we only use 180 images, \method still manages to learn useful correspondences.
This is a considerable advantage of our method over supervised previous work, as it can be trained from a small number of images and without human supervision. This allows the construction of general-purpose shape correspondence models for almost any category.

\section{Conclusion}%
\label{sec:conclusion}

We have introduced an unsupervised method to learn correspondence matching between a 3D template and images. Critically, this model can be trained without supervision and from less than 200 images, which makes it applicable to a vast number of objects.
This is a significant step beyond previous work that required lots of manually labelled correspondences.
We hope that \method will enable many downstream tasks where learnt robust correspondence estimation was previously impossible.

\paragraph{Ethics.}

We utilize the DensePose-LVIS dataset~\cite{neverova2020continuous} and PF-PASCAL~\cite{ham16proposal} for evaluation in a manner compatible with their terms.
The images may contain humans, but we only consider occurrences of animals and there is no processing of biometric information.
For further details on ethics, data protection, and copyright please see \url{https://www.robots.ox.ac.uk/~vedaldi/research/union/ethics.html}.

\paragraph{Acknowledgements.}

We thank Orest Kupyn for helpful discussions and Luke Melas-Kyriazi for proofreading this paper.
A. Shtedritski is supported by EPSRC EP/S024050/1.
A. Vedaldi is supported by ERC-CoG UNION 101001212.

\bibliographystyle{splncs04}
\bibliography{references,vedaldi_general,vedaldi_specific}

\clearpage
\begin{center}
	\textbf{\Large Appendix}
\end{center}

In this Appendix, we first discuss the limitations of our approach (\cref{sec:app-limitations}).
Then we discuss implementation details (\cref{sec:app-implementation}) and provide additional ablations (\cref{sec:app-ablations}).
Finally, we show more qualitative examples (\cref{sec:app-qualitative examples}) and show a failure mode we observe (\cref{sec:app-failure}).

\section{Limitations}%
\label{sec:app-limitations}

Our method has several limitations.
First, it relies on having a few hundred images per category, which might not always be possible for low-resource classes.
However, this is still a significant step forward from prior works, which need much more data and/or human annotations.

Next, the symmetry equivariance loss we propose assumes the shape is symmetric.
While this is true for all shapes we consider, there could be several instances where this assumption does not hold:
(i) if the shape is not symmetric by design, \eg, it an animal that misses a leg;
(ii) if the shape is articulated and thus not symmetric.
In that instance, the loss $\mathcal{L}_{eq}$ should not be used, which would lead to a small drop in performance.

Finally, our model only predicts image-to-vertex matching, whereas prior methods such as CSE~\cite{neverova2021discovering}
also predict segmentation masks.
However, prior methods do \emph{not} evaluate segmentation performance, as they are not competitive, and this is not the main point of the methods.
Furthermore, they use masks as an \emph{additional form of supervision}, as the model is additionally trained to predict masks, whereas we only use masks to sample points used during training (as not to try matching background points to the shape).

\section{Additional implementation details}%
\label{sec:app-implementation}

\subsection{Symmetry equivariance loss}

We automatically discover the plane of symmetry of the shape.
We assume the shape's plane of symmetry is either one of the $(x, y,  z)$ planes.
In practice, this is most often true.
We test each of the $(x, y,  z)$ planes as follows.
First, we center the mesh.
Then, for every plane, we mirror all vertices along that plane.
For every vertex, we find its nearest neighbour mirrored vertex.
We sum the Euclidean distances between all vertices and their mirrored nearest neighbours.
Intuitively, the correct plane of symmetry corresponds to the smallest sum of distances.
Finally, for every vertex $x$, we obtain its symmetric one $x_F$ by finding its nearest neighbour when we mirror the shape along the selected plane of symmetry.

\subsection{Training}

During training, we perform data augmentations: random crops, rotations, and colour jitters.
We perform these on both the natural and synthetic (generated with a depth-to-image model) images.
We train using the Adam optimizer for 40 epochs, using $lr=0.001$, which is decreased $10\times$ after 20 epochs.

\subsection{Synthetically generated ground-truth}

\begin{figure}[t]
\begin{center}
\includegraphics[width=1.0\textwidth]{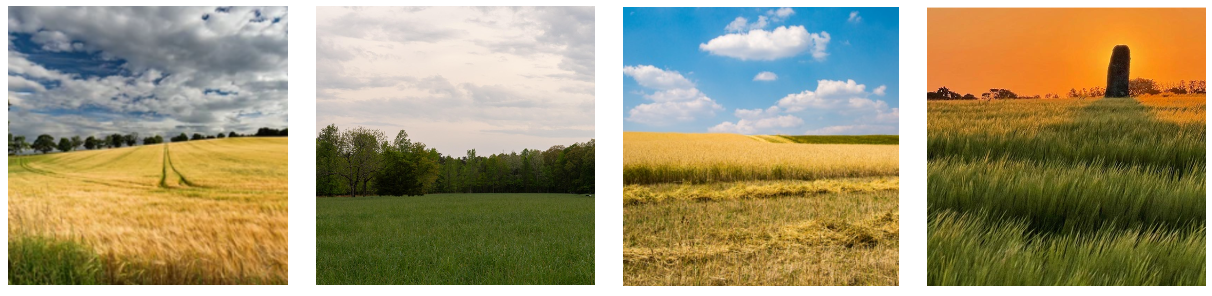}
\caption{\textbf{Background images.} To generate synthetic images, we sample from these, do a random crop, and predict depth.}\label{fig:app_background}
\end{center}
\end{figure}

\begin{figure}[t]
\begin{center}
\includegraphics[width=1.0\textwidth]{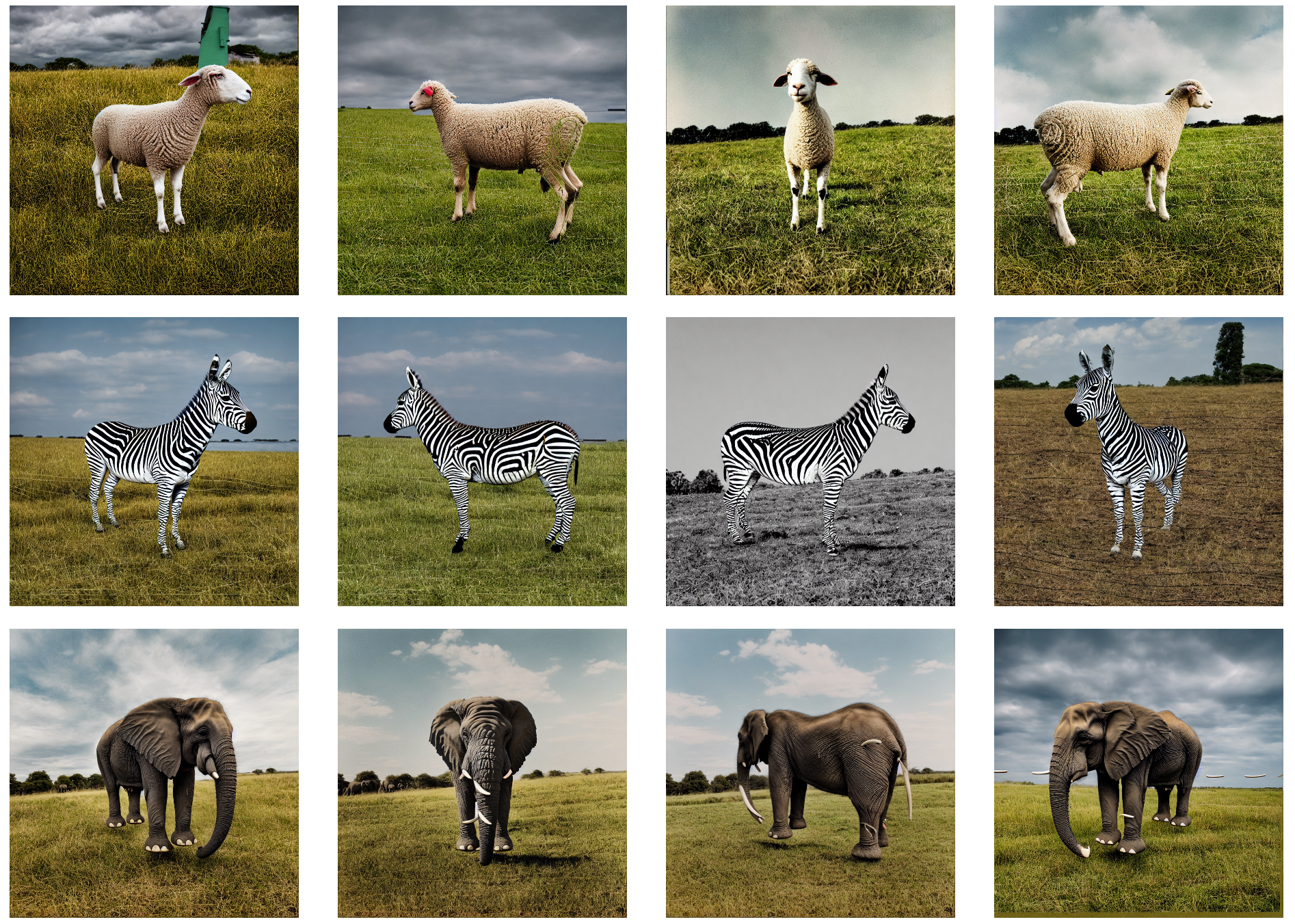}
\caption{\textbf{Synthetically generated images.}}\label{fig:synth}
\end{center}
\end{figure}

As discussed in the paper, to generate each synthetic image, we sample a viewpoint and a background.In practice, we sample from 4 background images (\cref{fig:app_background}), which we randomly crop before computing depth.We find that we can obtain diverse backgrounds with a small number of background templates by using different random seeds.
We sample viewpoints only from the side and front.We found that when we sample an image from the back, Stable Diffusion still tries to place a face on the back of the head, leading to unnatural-looking images.
We show more examples of generated images in \cref{fig:synth}.


\section{Additional ablations}%
\label{sec:app-ablations}

\subsection{Pseudo-ground truth}

We perform additional ablations on the features used to construct the pseudo-ground-truth $\Sigma$ in \cref{tab:appendix-ablations-data}.
First, we render shaded surfaces instead of surface normals and find that leads to a small drop in performance.
Next, we exclude the SD features from SD-DINO~\cite{zhang2023sd-dino}, and only use DINO features for matching.This makes computing the pseudo-ground-truth $\Sigma$ faster, as SD features are more expensive.As expected, we see decreased performance when only using DINO features.
\subsection{Number of training images}
We train our method using a different number of natural images in \cref{tab:appendix-ablations-num-data}.
We train on $\{50, 200, 500, \text{and } 2k+\}$ images, where $2k+$ is the number of images of the particular class in the dataset, falling between 2k and 3k.
We exclude the classes ``bear'' and ``sheep'' as they contain under $2k$ images.We see that with as few as 500 images, we achieve comparable performance to our full models.

\begin{table}[t]
\centering
\centering
\resizebox{0.65\textwidth}{!}{%
\begin{tabular}{llc}
\toprule
Ablation & & DensePose-LVIS \\
\midrule
& Ours & \textbf{24.9}\\
\midrule
Renders & \texttt{shaded} (instead of \texttt{normals})& 25.2 \\
\midrule
Features & w/o SD & 25.4 \\
\bottomrule
\end{tabular}}
\caption{\textbf{Data ablations.}
First, we ablate using shaded renders of the template shape instead of surface normals. Next, we train models using \emph{only} DINO features (w/o SD), as they are quicker to compute. We evaluate using geodesic distance (lower is better). }%
\label{tab:appendix-ablations-data}
\end{table}
\begin{table}[t]
\centering
\centering
\resizebox{0.45\textwidth}{!}{%
\begin{tabular}{lc}
\toprule
No\# images \; \; & DensePose-LVIS \\
\midrule
50 & 34.7 \\
200 & 29.8 \\
500 & 24.9 \\
2k+ & 22.3 \\
\bottomrule
\end{tabular}}
\caption{\textbf{Ablation of the number of training images.}
We ablate the number of training images used for each class. For this ablation, we exclude the ``bear'' and ``sheep'' classes as they have under 2k images, the other classes have between 2k and 3k images. We evaluate using geodesic distance (lower is better). }%
\label{tab:appendix-ablations-num-data}
\end{table}

\section{Qualitative examples}%
\label{sec:app-qualitative examples}

\begin{figure}[t]
\begin{center}
\includegraphics[width=0.99\textwidth]{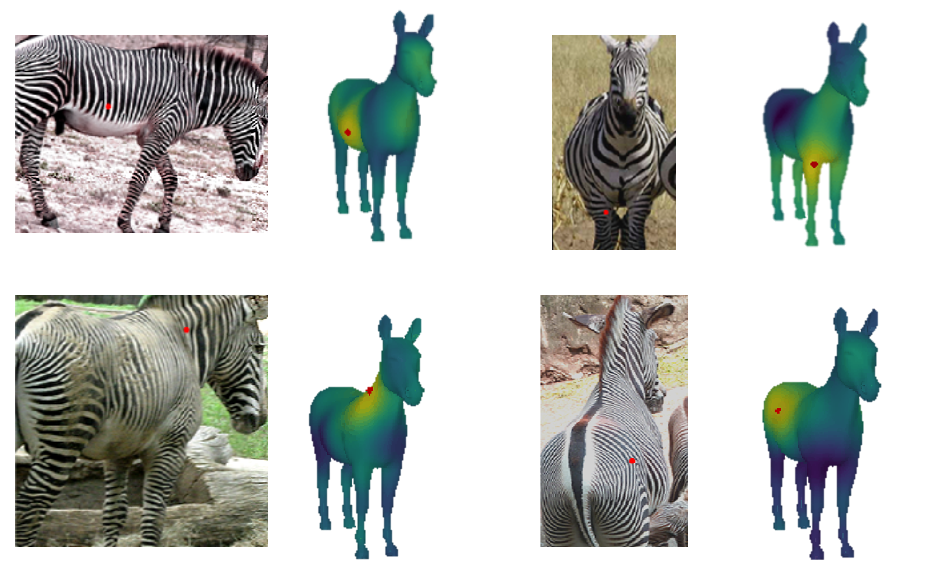}
\caption{\textbf{Similarity heatmaps.} We show similarity heatmaps between the visual feature sampled at the annotated location in red with the CSE embeddings learnt over the shape. We color every vertex according to that similarity and annotate the most similar vertex in red.}\label{fig:app_heatmaps}
\end{center}
\end{figure}

\begin{figure}[t]
\begin{center}
\includegraphics[width=0.99\textwidth]{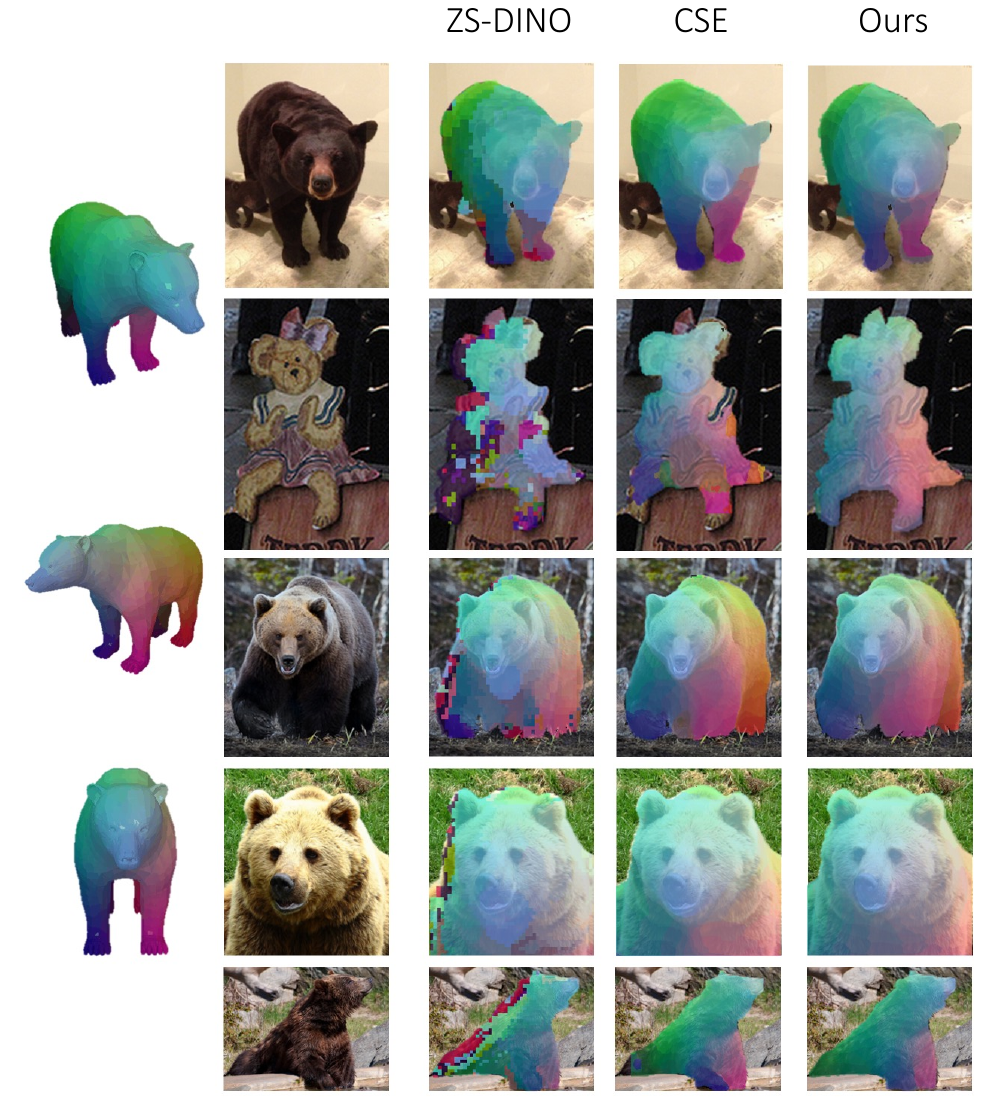}
\caption{\textbf{Qualitative results.}}\label{fig:app_bear}
\end{center}
\end{figure}

\begin{figure}[t]
\begin{center}
\includegraphics[width=0.99\textwidth]{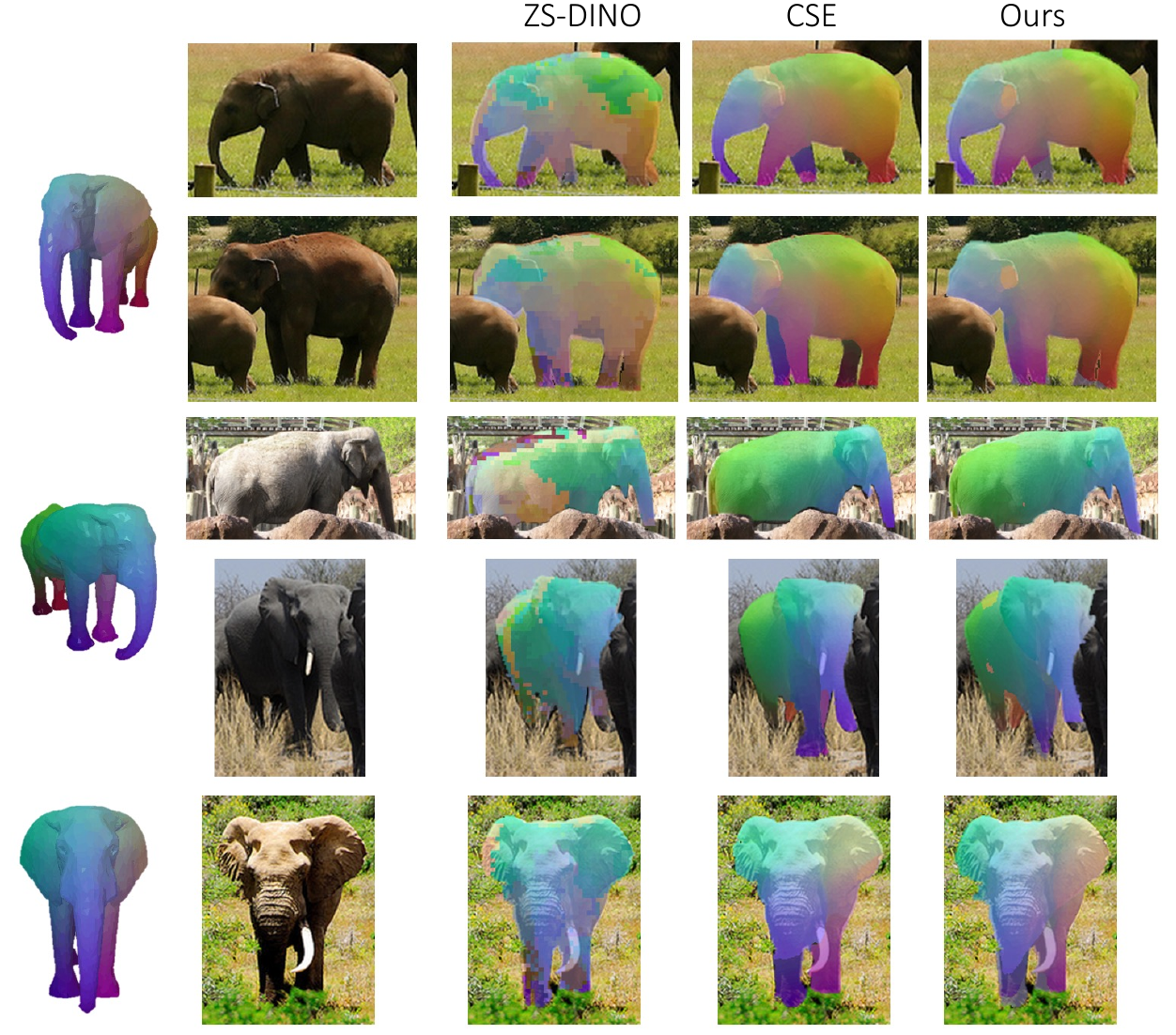}
\caption{\textbf{Qualitative results.}}\label{fig:app_elephant}
\end{center}
\end{figure}

\begin{figure}[t]
\begin{center}
\includegraphics[width=0.99\textwidth]{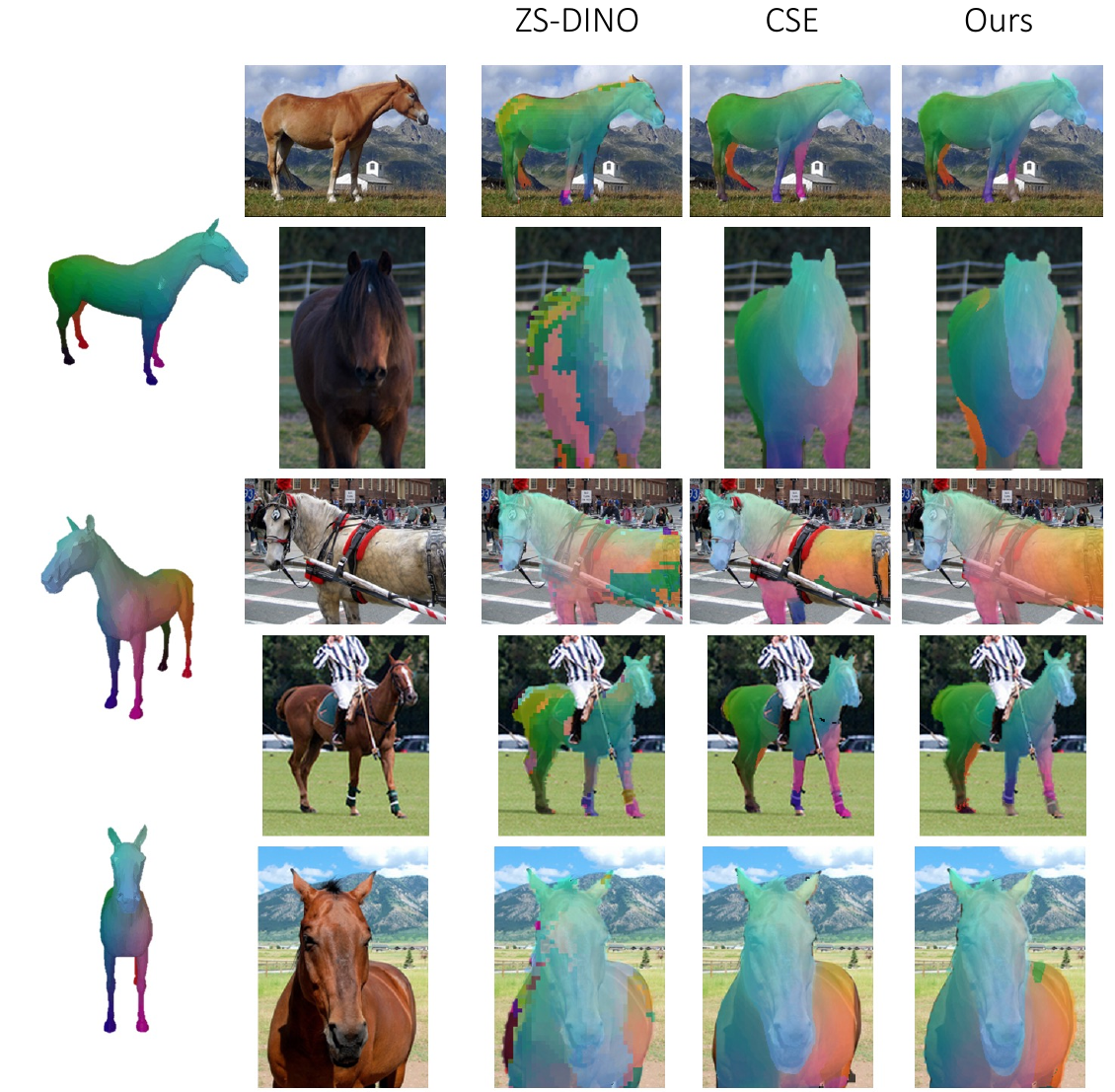}
\caption{\textbf{Qualitative results.}}\label{fig:app_horse}
\end{center}
\end{figure}

In \cref{fig:app_heatmaps} we show similarity heatmaps of the visual feature with the CSE embeddings over the shape. We show further qualitative examples of texture remapping in \cref{fig:app_bear,fig:app_elephant,fig:app_horse}.

\section{Failure case}%
\label{sec:app-failure}

\begin{figure}[t]
\begin{center}
\includegraphics[width=1.0\textwidth]{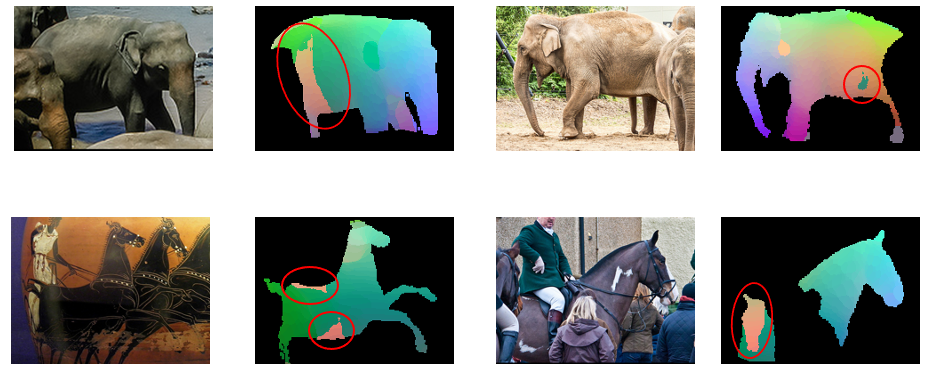}
\caption{\textbf{Failure cases.} We annotated failure cases in red, where the model predicts wrong patches.}\label{fig:failure}
\end{center}
\end{figure}

We observe a failure case, where the model predicts wrong patches (\cref{fig:failure}). We notice that these patches correspond to the same semantic part, but on opposite sides (\eg, a patch of ``left belly'' is predicted where there should be ``right belly'').

\end{document}